\pdfoutput=1

\documentclass[11pt]{article}

\usepackage[]{ACL2023}
\usepackage{times}
\usepackage{latexsym}
\usepackage[normalem]{ulem}
\usepackage[T1]{fontenc}
\usepackage[utf8]{inputenc}
\usepackage{microtype}
\usepackage{inconsolata}
\usepackage{graphicx}
\usepackage{tabularx}
\usepackage{booktabs}
\usepackage{makecell}

\title{GAIA Search: Hugging Face and Pyserini \\ Interoperability for NLP Training Data Exploration
}

\newcommand{\hf}{$^1$}
\newcommand{\sapienza}{$^2$}
\newcommand{\hfsapienza}{$^{1,2}$}
\newcommand{\uw}{$^3$}
\newcommand{\lu}{$^4$}
\newcommand{\scad}{$^5$}
\newcommand{\luscad}{$^{4,5}$}
\newcommand{\eleuther}{$^6$}

\author{
Aleksandra Piktus\hfsapienza{}
Odunayo Ogundepo\uw{}
Christopher Akiki\luscad{} 
Akintunde Oladipo\uw{}\\
\textbf{
Xinyu Zhang\uw{} 
Hailey Schoelkopf\eleuther{}
Stella Biderman$^{6,7}$
Martin Potthast\luscad{}
Jimmy Lin\uw{}} \\
\\
\hf{}Hugging Face \ \sapienza{}Sapienza University 
\uw{}University of Waterloo \\ \lu{}Leipzig University \ \scad{}ScaDS.AI \ \eleuther{} EleutherAI \\ ${}^7$Booz Allen Hamilton\\
{\tt piktus@huggingface.co}
\\
}

\begin{document}
\maketitle
\begin{abstract}
Noticing the urgent need to provide tools for fast and user-friendly qualitative analysis of large-scale textual corpora of the modern NLP, we propose to turn to the mature and well-tested methods from the domain of Information Retrieval (IR)---a research field with a long history of tackling TB-scale document collections. We discuss how Pyserini---a widely used toolkit for reproducible IR research can be integrated with the Hugging Face ecosystem of open-source AI libraries and artifacts. We leverage the existing functionalities of both platforms while proposing novel features further facilitating their integration. Our goal is to give NLP researchers tools that will allow them to develop retrieval-based instrumentation for their data analytics needs with ease and agility.
We include a Jupyter Notebook-based walk through the core interoperability features, available \href{https://github.com/huggingface/gaia}{on GitHub}.
We then demonstrate how the ideas we present can be operationalized to create a powerful tool for qualitative data analysis in NLP. We present GAIA Search---a search engine built following previously laid out principles, giving access to four popular large-scale text collections. GAIA serves a dual purpose of illustrating the potential of methodologies we discuss but also as a standalone qualitative analysis tool that can be leveraged by NLP researchers aiming to understand datasets prior to using them in training. GAIA is hosted live on \href
{https://huggingface.co/spaces/spacerini/gaia}{Hugging Face Spaces}.

\end{abstract}

\section{Introduction}
Training large language models, or LLMs~\cite{NEURIPS2020_1457c0d6,J1WhitePaper,Rae2021ScalingLM,https://doi.org/10.48550/arxiv.2201.11990,bloom,https://doi.org/10.48550/arxiv.2204.02311, touvron2023llama}, established itself as the central task of the modern Natural Language Processing (NLP) research. The attempts to understand the scaling laws of LLMs led researchers to believe that simply increasing the number of parameters may not bring the desired improvements without a simultaneous increase in the size of the LLM training data~\cite{kaplan2020scaling,https://doi.org/10.48550/arxiv.2203.15556}.
These observations only increased an already pressing need for massive textual datasets, fueling the proliferation of Web-based corpora of TB-scale
created with varying levels of curation and quality control.

\begin{figure*}[!t]
\includegraphics[width=\linewidth]{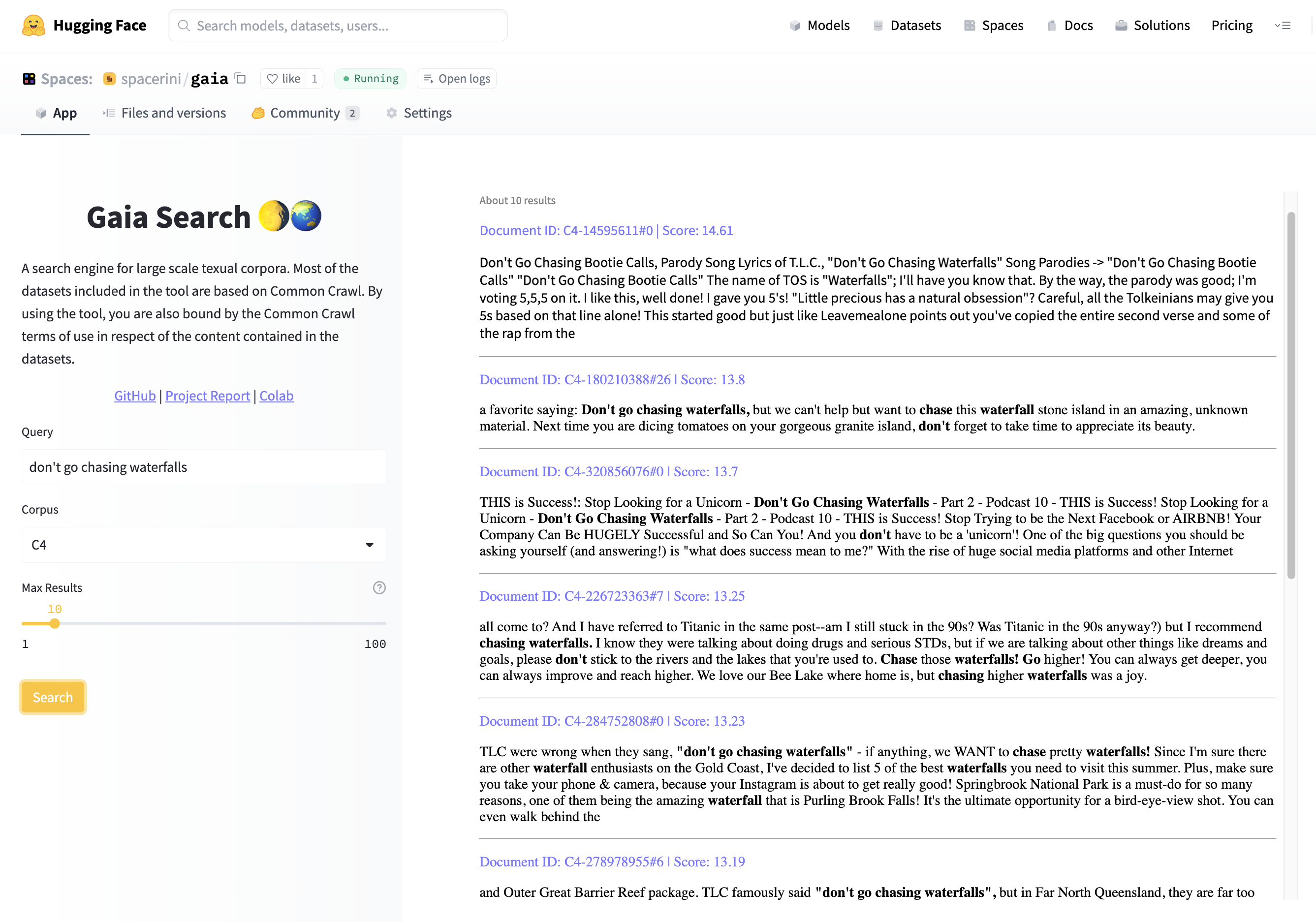}
\caption{The user interface of GAIA Search.}
\label{fig:ui}
\end{figure*}

Rather than investing in scraping the Web on their own, dataset creators typically turn to Common~Crawl\footnote{\url{https://commoncrawl.org/}} as the main source of text to include in their corpora. A repository of Web snapshots dating back to 2011, Common Crawl contains various types of low-quality text~\cite{luccioni-viviano-2021-whats}. Pre-processing steps commonly introduced by dataset creators aiming to filter out undesired content include removing any documents with words matching a pre-defined, static blacklist, like in the case of C4~\cite{10.5555/3455716.3455856}, perplexity-based filtering like in CCNet and ROOTS~\cite{https://doi.org/10.48550/arxiv.1911.00359,laurencon2022the}, removing malformed text via simple text statistics like in the case of OSCAR~\cite{abadji-etal-2022-towards} or through deduplication, studied extensively by~\citet{lee-etal-2022-deduplicating}.
However, the generated artifacts still tend to contain a multitude of worrying phenomena, such as synthetic data \cite{DodgeSapEtAl_2021_Documenting_Large_Webtext_Corpora_Case_Study_on_Colossal_Clean_Crawled_Corpus}, private and copyrighted data \cite{HuangShaoEtAl_2022_Are_Large_PreTrained_Language_Models_Leaking_Your_Personal_Information} or incorrect language codes and translations \cite{KreutzerCaswellEtAl_2022_Quality_at_Glance_Audit_of_WebCrawled_Multilingual_Datasets}. A lack of representation of diversity and socio-cultural and socio-economic biases constitute another big challenge of Common Crawl and datasets derived from it \cite{10.1145/3442188.3445922,BlodgettBarocasEtAl_2020_Language_Technology_is_Power_Critical_Survey_of_Bias_in_NLP,FieldBlodgettEtAl_2021_Survey_of_Race_Racism_and_AntiRacism_in_NLP,StanczakAugenstein_2021_Survey_on_Gender_Bias_in_Natural_Language_Processing,beaulieu-leonelli-2021-data}.

Aware of the mounting problems with training data for modern LLMs, and appreciating the value of data exploration for better modeling in general, we focus our current work on building tools that can facilitate the qualitative analysis of NLP datasets. We propose to leverage the extensive experience of the Information Retrieval community in building relevance-based search indices for large-scale document collections and put it into practice in the context of NLP data exploration work. We follow with a demonstration of ways in which the interoperability between Pyserini~\cite{Lin_etal_SIGIR2021_Pyserini}, a leading toolkit for reproducible IR research on one side, and Hugging Face\footnote{\url{https://huggingface.co/}}, a platform for open AI research on the other, can be leveraged to build tools for easy and effective analysis of textual data. To facilitate the adoption of the proposed methods we provide a collection of Jupyter Notebooks with step-by-step explanations of explored functionalities available \href{https://github.com/huggingface/gaia/tree/main/notebooks}{on GitHub}.

Finally, we release GAIA---a simple, yet powerful search engine giving relevance-based interface to four popular, large-scale, textual datasets, namely C4~\cite{10.5555/3455716.3455856}, the Pile~\cite{https://doi.org/10.48550/arxiv.2101.00027,pile2022datasheet}, ROOTS~\cite{laurencon2022the} and captions from LAION-2B-en~\cite{https://doi.org/10.48550/arxiv.2210.08402}.   All considered datasets rely to a big extent on data mined from Common Crawl. GAIA benefits from the interoperability between Pyserini and Hugging Face that we discuss in the first part of the paper, while also constituting a standalone contribution which can benefit the NLP research community by making it easy to study leading corpora qualitatively. GAIA is available online at \url{hf.co/spaces/spacerini/gaia}.

\begin{table*}[t!]
\footnotesize
\centering
\begin{tabular}{lllrrrr}
\toprule
 Dataset & Reference & Hugging Face Hub link &  \# docs  & \# snippets &  Data Size & Index Size \\
 \midrule
C4 & \citet{10.5555/3455716.3455856}	& \href{https://huggingface.co/datasets/c4}{\texttt{c4}} & 365M & 1,587M & 829GB &  1.3TB \\
The Pile & \citet{https://doi.org/10.48550/arxiv.2101.00027} & 
 \href{https://huggingface.co/datasets/EleutherAI/the_pile_deduplicated}{\texttt{the\_pile\_deduplicated}} & 134M & 673M & 825GB  &1.2TB \\
ROOTS	& \citet{laurencon2022the} & \href{https://huggingface.co/bigscience-data}{\texttt{bigscience-data}}	& 598M & 2,171M &  1.6TB & 2.6TB \\
LAION	& \citet{https://doi.org/10.48550/arxiv.2210.08402}	& \href{https://huggingface.co/datasets/laion/laion2B-en}{\texttt{laion2B-en}} & 2,322M	& 1,351M & 503GB &446GB \\
\midrule
\multicolumn{3}{r}{Total} &  3,419M & 5,782M  &  3.76 TB & 5.55TB \\
\bottomrule
\end{tabular}
\caption{Datasets included in the GAIA Search tool. All numbers refer to the size of the \texttt{train} split of the data.}
\label{tab:gaia_details}
\end{table*}

\section{Background}
The ability to analyze large collections of textual data is core in multiple research and engineering disciplines. While the industrial standard is to rely on robust, scalable database and data analytics infrastructure, in the research environment, we typically resort to more local, granular and flexible, if ad-hoc, solutions which leverage toolkits such as NumPy~\cite{harris2020array}, Pandas~\cite{reback2020pandas,pandas-paper}, SciPy~\cite{2020SciPy-NMeth} and others. A common research approach to data analytics involves using one of the aforementioned packages in combination with Jupyter Notebooks\footnote{\url{https://jupyter.org/}}. Notebooks make it easy to deploy and share analyses, however, typically they remain essentially non-interactive, requiring at least a basic understanding of programming to be able to work with them efficiently. With the commodification of AI, and NLP in particular, and the expansion of NLP technologies into research areas beyond AI~\cite{https://doi.org/10.48550/arxiv.2203.03540,10.1093/alh/ajv029,8029924,cancer-nlp-no-code}, the need for easy to use, no-code tools for understanding AI artifacts arises. This need is partly addressed by Python packages such as Streamlit\footnote{\url{https://streamlit.io/}} and Gradio\footnote{\url{https://gradio.app/}}, designed to facilitate the creation of interactive Machine Learning (ML) demos. As the authors of the Gradio white paper~\cite{https://doi.org/10.48550/arxiv.1906.02569} point out, the accessibility and ease of use of the analysis tools is critical if we want to build an understanding of AI and trust in it. The Hugging Face Spaces platform, providing free hosting of both Streamlit, Gradio, and Docker-based applications, serves this exact purpose. However, it puts emphasis on demonstrating the capabilities of models while paying less attention to the datasets used to train them.

Even more so than in NLP, the evaluation of IR systems is heavily dependent on the implementation details of the retrieval systems serving the search indices being evaluated. The lack of standardisation of IR evaluation was the main motivator behind creating Anserini~\cite{10.1145/3077136.3080721}, a Lucene\footnote{\url{https://lucene.apache.org/}}-based toolkit for reproducible IR research, and the follow-up Pyserini~\cite{Lin_etal_SIGIR2021_Pyserini}---a convenient Python API to the underlying Java-based implementation of Anserini. While it is relatively easy to build and serve search indices backed by Pyserini and Lucene, the task of building and deploying interactive user interfaces generally comes with a higher engineering barrier of entry.

Relevance-based search interfaces have been previously explored in the context of NLP---e.g. in the C4 analysis~\cite{DodgeSapEtAl_2021_Documenting_Large_Webtext_Corpora_Case_Study_on_Colossal_Clean_Crawled_Corpus}, in COVID-related datasets~\citep{zhang2020covidex} or in news quotes~\cite{Vukovi__2022}. Rather than focusing only on providing finished artifacts, however, we intend our current work to serve as a reference and inspiration for NLP researchers looking to develop and deploy similar applications by themselves.

We attempt to bring together the power of Pyserini-backed retrieval and the agility of ML demo development within the Hugging Face ecosystem to serve the goal of building intuitive data exploration tools. We believe that resulting applications will make a great difference for NLP researchers trying to study their data qualitatively, as well as to non-technical researchers looking for tools allowing them to perform dataset analysis in a no-code fashion. We propose our search engine GAIA as a compelling case in point. 

\section{Pyserini and Hugging Face: From Data to Search}
\label{sec:interop}
In the current section we discuss core components which need to be considered when building a search application for textual datasets. We focus on how each step can be facilitated by the use of Pyserini, Hugging Face, or a combination of the two. We also provide hands-on tutorials covering basic concepts and search engine building blocks such as \href{https://nbviewer.org/github/huggingface/gaia/blob/main/notebooks/00-indexing.ipynb}{data loading and indexing}, \href{https://nbviewer.org/github/huggingface/gaia/blob/main/notebooks/01-tokenization.ipynb}{tokenization}, \href{https://nbviewer.org/github/huggingface/gaia/blob/main/notebooks/02-searching.ipynb}{search}, and \href{https://nbviewer.org/github/huggingface/gaia/blob/main/notebooks/03-analysis.ipynb}{index analysis}. We further release the \href{https://github.com/huggingface/gaia/tree/main/preprocessing}{pre-processing}, \href{https://github.com/huggingface/gaia/tree/main/web}{backend} and \href{https://huggingface.co/spaces/spacerini/gaia/blob/main/app.py}{frontend code} that allowed us to index 3.5 billion documents---chunked into 5.8 billion snippets---and serve 5.55TB worth of BM25 indexes.

\subsection{Data Access}
The Hugging Face hub is the repository of over 20,000 datasets from across AI domains. This includes the most popular large-scale text corpora in NLP---for example all the datasets we consider in GAIA (see Table~\ref{tab:gaia_details} for details), but also other popular large scale text datasets such as \href{https://hf.co/datasets/oscar-corpus/OSCAR-2201}{OSCAR}~\cite{abadji-etal-2022-towards} and \href{https://hf.co/datasets/bigcode/the-stack}{The Stack}~\cite{https://doi.org/10.48550/arxiv.2211.15533} among many others. Each dataset hosted on the Hub can be accessed locally using the \texttt{datasets}~\citep{lhoest-etal-2021-datasets} library which provides convenient and parallelizable APIs for downloading and processing the data. Memory-mapping is supported by default and uses the efficient an Apache Arrow format,\footnote{\url{https://arrow.apache.org/}} making it possible to seamlessly handle datasets surpassing the RAM constraints of a given machine. \texttt{Datasets} also provide a streaming functionality which dispenses of downloading data to disk, making it possible to work with larger-than-disk datasets.

\subsection{Tokenization}
Tokenization is a crucial pre-processing step in NLP and Information Retrieval.
In the context of IR, this process typically includes removing stop words, stemming, lemmatization, and removing non-alphanumeric characters. By default, Pyserini uses Lucene analyzers---heuristics-based algorithms designed for various languages and use cases, to tokenize text. The drawback of this approch is that only some languages have dedicated analyzers, while others have to resort to simply breaking on whitespace, which inadvertently leads to suboptimal performance.

An alternative to whitespace tokenization that has shown promise in Information Retrieval and is a mainstay in NLP is subword tokenization~\citep{Mielke2021BetweenWA}, a process which splits words into smaller units based on their frequency in the corpus. Hugging Face provides a range of tokenizers that are specifically designed to work with its pre-trained transformer language models, as well as the means to train such tokenizers~\citep{anthony_moi_2022_tokenizers}.

As of recently, Pyserini can leverage Hugging Face pre-trained subword tokenizers to improve indexing and searching for multiple languages. Pre-trained tokenizers from Hugging Face can serve as drop-in replacements for Lucene Analyzers, improving retrieval effectiveness, particularly in low-resource languages~\cite{https://doi.org/10.48550/arxiv.2210.05481}. This interoperability between Hugging Face and Pyserini makes it easy for researchers to incorporate deep learning-based language models into their information retrieval workflows and opens up new avenues for research in the field.

\subsection{Building the Index}
Indexing constitutes the core functionality of Pyserini. The library enables experiments with bag-of-words sparse retrieval using Lucene, and dense vector retrieval using \texttt{Faiss}~\citep{johnson2019billion-faiss}, as well as hybrid retrieval combining the two. Though this project focuses solely on sparse retrieval using BM25 indexes, Pyserini's dense encoding and retrieval API would make it very easy to adapt all examples and demos to this paradigm.

\paragraph{Offline Indexing.} Arrow-backed Hugging Face datasets readily lend themselves to being indexed by Pyserini's standard Lucene indexer. In principle, one can build an index of a Hugging Face dataset simply by downloading it locally and then passing the file path to the Pyserini indexer via a command line argument. The scenario where a pre-processing step is required in between the data download and the indexing step---as with document segmentation which we discuss later in Section~\ref{sec:gaia}--can be realised straightforwardly for smaller datasets, which fit both on disk and into RAM. The larger-than-RAM datasets which fit on disk, can be easily sharded into any of the disk text formats supported by Pyserini (those include CSV, TSV, JSON, and JSONL) and processed concurrently within RAM limits to be then passed to the indexer.

\paragraph{Datasets Streaming.} As of recently, it is also possible to index datasets which don't fit on disk.\footnote{Note however, that the resulting index does have to fit on disk. As a result, we envision this functionality to be particularly convenient for scenarios where either the dataset or the index may be able to fit on disk, but both do not---a common scenario when dealing with TB-scale artefacts.} This new addition to Pyserini---one that resulted out of our current collaboration---allows users to stream text into the index directly---in other words, build an index on the fly from a text stream rather than from a static file saved on disk. As a result, larger-than-disk collections can be streamed from the Hugging Face Hub directly into the local indexing process. Data streaming can also improve experimental agility for smaller datasets, by removing the data downloads step from the Hugging Face dataset---Pyserini index pipeline.

\subsection{Backend: Custom Pyserini Server}
Once the data index is ready we need a way to host it and serve the search functionality to the clients. We propose a simple Python-based, Pyserini server implementation for GAIA, which can be easily generalized to other use-cases. The server code can be accessed \href{https://github.com/huggingface/gaia/tree/main/web}{on GitHub}.

\subsection{Frontend: Interactive Demos}
Providing interactive demos which enable the exploration of AI artifacts is crucial in order to be able to collaborate across research disciplines and share results with colleagues without imposing the burden of setting up their own engineering stack on them. By offering the hosting of Gradio and Streamlit applications Hugging Face Spaces meet this need perfectly. We encourage readers to follow the implementations of GAIA for an example of how to build a simple UI for a search tool.

\section{Case Study: GAIA Search}
\label{sec:gaia}
Relevance-based search tools have the potential of the largest impact on massive-scale datasets, common in modern NLP. Unlike with smaller data collections, where simpler investigation strategies, e.g. via a combination of Pandas and Jupyter Notebooks, may be feasible, huge datasets are generally too cumbersome to process this way. A big benefit of search engines in the form that we propose is also the fact that after being set up, they require no engineering skills or extensive computing resources to operate, expanding the community of potential users. We demonstrate this with GAIA search, available online at \url{hf.co/spaces/spacerini/gaia}.

\subsection{Included Datasets}
GAIA proposes a simple interface to four large-scale textual datasets---C4, The Pile, ROOTS, and captions from LAION-2B-en.
The reader may consult Table~\ref{tab:gaia_details} for details on respective datasets.
All of the datasets included in GAIA are sourced at least partly from Common Crawl. The users of the tool are therefore bound by the Common Crawl terms of use\footnote{\url{https://commoncrawl.org/terms-of-use/}} in respect of the content contained in the datasets. Additionally, in order to respect the data subjects' rights~\cite{10.1145/3531146.3534637} we refrain from presenting full documents in the tool, and instead include snippets of at most 256 words. We redact the personally identifiable information (PII) on all search results on the backend side, using the PII redaction script open-sourced alongside the BigScience\footnote{\url{bigscience.huggingface.co}} language model BLOOM~\cite{bloom}.
Below we discuss details of the respective datasets' pre-processing.

\paragraph{C4.} This is a dataset fully sourced from Common Crawl. We index the variant of the English split of the dataset \href{https://huggingface.co/datasets/c4}{available on the Hugging Face hub}. C4 has been used to train T5~\cite{10.5555/3455716.3455856}, a major encoder-decoder model with a multitude of downstream applications, parts of it have also contributed to the training of other LLMs, e.g. LaMDA~\cite{https://doi.org/10.48550/arxiv.2201.08239} and Chinchilla~\cite{https://doi.org/10.48550/arxiv.2203.15556}, which makes it a compelling dataset to study.

\paragraph{The Pile.}
This corpus has been a standard dataset for many English LLM releases from various organizations \cite{biderman2023pythia, black2021gpt, gpt-j, black2022gpt, https://doi.org/10.48550/arxiv.2201.11990, WuDao, https://doi.org/10.48550/arxiv.2205.01068, J1WhitePaper}, so we believe that it is important to expose its contents to public view. 
The Pile is an English-only corpus containing multiple sub-corpora from various sources \citep{pile2022datasheet}. We use a variant of The Pile which has been deduplicated with MinhashLSH and a threshold of 0.87, following the advice of \citet{lee-etal-2022-deduplicating}. Notably, this variant of the Pile has also been used to train an LLMs~\cite{biderman2023pythia}. We hope that providing the search interface will allow further investigation of the subjective differences between deduplicated and unprocessed corpora. Both \href{https://huggingface.co/datasets/the_pile}{the canonical variant of The Pile} and it's \href{https://huggingface.co/datasets/EleutherAI/the_pile_deduplicated}{deduplicated counterpart} are available on the Hugging Face Hub.

\paragraph{ROOTS.} Developed for the purpose of training BLOOM~\cite{bloom}, this is the only multilingual dataset available in GAIA. We therefore, create independent indices for each language or language group provided in the corpus, resulting in 13 separate indices---Arabic, Catalan, Code (comprising all programming languages included in the corpus), English, Spanish, Basque, French, Indonesian, Indic and Niger-Congo (language groups), Portuguese, Vietnamese and Chinese. We return results for each index when issuing queries in the tool.

\paragraph{LAION-2B-en} LAION is a dataset of image caption---image URL pairs scraped from the Web. It has been used to train Stable Diffusion~\cite{rombach2021highresolution}, a textual-prompt-based image generation model, constituting an open-source counterpart to OpenAI's DALL·E~2~\cite{https://doi.org/10.48550/arxiv.2204.06125}.
We use LAION-2B-en, the subset of the original dataset with captions in English, as the starting point for further pre-processing. We start by deduplicating captions, which yields clusters of image URLs with identical captions (deduplication code is available \href{https://github.com/huggingface/gaia/tree/main/preprocessing}{on GitHub}). We then index unique captions. For textual queries to our tool, we return results consisting of the relevant captions. Alongside each result, we include the list of associated image URLs.

\subsection{Implementation and Functionality}
The implementation of GAIA makes use of a variety of interoperability features we've discussed in Section~\ref{sec:interop}.
As detailed in Table~\ref{tab:gaia_details}, all of the considered datasets are available on the Hugging Face Hub. We download and segment them locally. Such segmented datasets are then provided as input to a Pyserini indexer. We leverage Streamlit to build the user interface for our tool and host it on Hugging Face Spaces. On the backend side, the indices are served from Hugging Face provisioned machines. We open-source helper functions for segmenting long documents and the backend server code at \url{github.com/huggingface/gaia}.

\section{Limitations and Future Plans}
\label{sec:limitations}
A major area for consideration when developing data access tools is that of data governance, privacy and data ownership~\cite{10.1145/3531146.3534637,CarliniTramerEtAl_2020_Extracting_Training_Data_from_Large_Language_Models}. In our current work we focus on the technical aspects of giving access to large data collections, however, we urge users to consider data governance principles when designing their own tools.
In terms of the infrastructure, the cost and complexity of hosting the retrieval index falls on the creator of the tool, which can be easy to manage for small datasets but becomes more problematic when entering the realm of TB-scale corpora. We are currently investigating a parallel workstream that could address this limitation at least partly.

\section{Conclusions}
We showcase interoperability between Hugging Face and Pyserini and provide value to the NLP community by demonstrating easy ways to perform high-quality, large-scale retrieval with open-source tools. We also introduce GAIA - a search engine for retrieval-based exploration of four major textual datasets. We wish to encourage NLP and IR practitioners to follow our examples and build their own tools to explore both large and smaller-scale textual datasets.

\section{Acknowledgements}
Authors would like to thank Carlos Muñoz Ferrandis, 
Daniel van Strien, Katie Link and Quentin Lhoest for valuable tips and suggestions.
This research was also supported in part by the Natural Sciences and Engineering Research Council (NSERC) of Canada.

\section{Impact Statement}
As mentioned in Section~\ref{sec:limitations}, accessing large-scale, web-scraped textual corpora comes with a variety of ethical considerations, pertaining to the protection of rights of the data owners and people whose privacy or copyright might be infringed upon. We introduce guardrails, namely the PII redaction and the segmentation of documents into short snippets, preventing the ability to reconstruct full documents or full corpora, into the GAIA Search design. We strongly encourage researchers aiming to build similar tools to do the same. Overall, a lot of these problems seem to occur because we're proposing the tool only after the datasets have been created and models trained on them. The workflow we envision for future research projects would involve building data exploration tools prior to the release of the datasets, so that core problems can be observed, studied and addressed before datasets reach an external audience.

\bibliography{anthology,custom}

\begin{thebibliography}{56}
\expandafter\ifx\csname natexlab\endcsname\relax\def\natexlab#1{#1}\fi

\bibitem[{Abadji et~al.(2022)Abadji, Ortiz~Suarez, Romary, and
  Sagot}]{abadji-etal-2022-towards}
Julien Abadji, Pedro Ortiz~Suarez, Laurent Romary, and Beno{\^\i}t Sagot. 2022.
\newblock \href {https://aclanthology.org/2022.lrec-1.463} {Towards a cleaner
  document-oriented multilingual crawled corpus}.
\newblock In \emph{Proceedings of the Thirteenth Language Resources and
  Evaluation Conference}, pages 4344--4355, Marseille, France. European
  Language Resources Association.

\bibitem[{Abid et~al.(2019)Abid, Abdalla, Abid, Khan, Alfozan, and
  Zou}]{https://doi.org/10.48550/arxiv.1906.02569}
Abubakar Abid, Ali Abdalla, Ali Abid, Dawood Khan, Abdulrahman Alfozan, and
  James Zou. 2019.
\newblock \href {https://doi.org/10.48550/ARXIV.1906.02569} {Gradio:
  Hassle-free sharing and testing of ml models in the wild}.

\bibitem[{Beaulieu and Leonelli(2021)}]{beaulieu-leonelli-2021-data}
Anne Beaulieu and Sabina Leonelli. 2021.
\newblock \emph{Data and Society: A Critical Introduction}.
\newblock Sage.

\bibitem[{Bender et~al.()Bender, Gebru, McMillan-Major, and
  Shmitchell}]{10.1145/3442188.3445922}
Emily~M. Bender, Timnit Gebru, Angelina McMillan-Major, and Shmargaret
  Shmitchell.
\newblock On the dangers of stochastic parrots: Can language models be too
  big?, year = {2021}, isbn = {9781450383097}, publisher = {Association for
  Computing Machinery}, address = {New York, NY, USA}, url =
  {https://doi.org/10.1145/3442188.3445922}, doi = {10.1145/3442188.3445922},
  abstract = {The past 3 years of work in NLP have been characterized by the
  development and deployment of ever larger language models, especially for
  English. BERT, its variants, GPT-2/3, and others, most recently Switch-C,
  have pushed the boundaries of the possible both through architectural
  innovations and through sheer size. Using these pretrained models and the
  methodology of fine-tuning them for specific tasks, researchers have extended
  the state of the art on a wide array of tasks as measured by leaderboards on
  specific benchmarks for English. In this paper, we take a step back and ask:
  How big is too big? What are the possible risks associated with this
  technology and what paths are available for mitigating those risks? We
  provide recommendations including weighing the environmental and financial
  costs first, investing resources into curating and carefully documenting
  datasets rather than ingesting everything on the web, carrying out
  pre-development exercises evaluating how the planned approach fits into
  research and development goals and supports stakeholder values, and
  encouraging research directions beyond ever larger language models.},
  booktitle = {Proceedings of the 2021 ACM Conference on Fairness,
  Accountability, and Transparency}, pages = {610–623}, numpages = {14},
  location = {Virtual Event, Canada}, series = {FAccT '21}.

\bibitem[{Bhardwaj et~al.(2017)Bhardwaj, Nambiar, and Dutta}]{8029924}
Rohan Bhardwaj, Ankita~R. Nambiar, and Debojyoti Dutta. 2017.
\newblock \href {https://doi.org/10.1109/COMPSAC.2017.164} {A study of machine
  learning in healthcare}.
\newblock In \emph{2017 IEEE 41st Annual Computer Software and Applications
  Conference (COMPSAC)}, volume~2, pages 236--241.

\bibitem[{Biderman et~al.(2022)Biderman, Bicheno, and Gao}]{pile2022datasheet}
Stella Biderman, Kieran Bicheno, and Leo Gao. 2022.
\newblock \href {http://arxiv.org/abs/2201.07311} {Datasheet for the pile}.
\newblock \emph{CoRR}, abs/2201.07311.

\bibitem[{Biderman et~al.(2023)Biderman, Schoelkopf, Anthony, Bradley, O'Brien,
  Hallahan, Khan, Purohit, Prashanth, Skowron, Sutawika, and van~der
  Wal}]{biderman2023pythia}
Stella Biderman, Hailey Schoelkopf, Quentin Anthony, Herbie Bradley, Kyle
  O'Brien, Eric Hallahan, Mohammad~Aflah Khan, Shivanshu Purohit, USVSN~Sai
  Prashanth, Aviya Skowron, Lintang Sutawika, and Oskar van~der Wal. 2023.
\newblock \href {https://doi.org/10.48550/arXiv.2201.07311} {Pythia: a scaling
  suite for language model interpretability research}.
\newblock \emph{Computing Research Repository}.
\newblock Version 1.

\bibitem[{Black et~al.(2021)Black, Gao, Wang, Leahy, and
  Biderman}]{black2021gpt}
Sid Black, Leo Gao, Phil Wang, Connor Leahy, and Stella Biderman. 2021.
\newblock \href {https://www.github.com/eleutherai/gpt-neo} {{GPT-Neo}: Large
  scale autoregressive language modeling with {Mesh-TensorFlow}}.
\newblock \emph{GitHub}.

\bibitem[{Black et~al.(2022)Black, Biderman, Hallahan, Anthony, Gao, Golding,
  He, Leahy, McDonell, Phang et~al.}]{black2022gpt}
Sidney Black, Stella Biderman, Eric Hallahan, Quentin Anthony, Leo Gao,
  Laurence Golding, Horace He, Connor Leahy, Kyle McDonell, Jason Phang, et~al.
  2022.
\newblock {GPT-NeoX-20B}: An open-source autoregressive language model.
\newblock In \emph{Proceedings of BigScience Episode \#5--Workshop on
  Challenges \& Perspectives in Creating Large Language Models}, pages 95--136.

\bibitem[{Blodgett et~al.(2020)Blodgett, Barocas, Daum{\'e}~III, and
  Wallach}]{BlodgettBarocasEtAl_2020_Language_Technology_is_Power_Critical_Survey_of_Bias_in_NLP}
Su~Lin Blodgett, Solon Barocas, Hal Daum{\'e}~III, and Hanna Wallach. 2020.
\newblock \href {https://doi.org/10.18653/v1/2020.acl-main.485} {Language
  ({{Technology}}) is {{Power}}: {{A Critical Survey}} of ``{{Bias}}'' in
  {{NLP}}}.
\newblock In \emph{Proceedings of the 58th {{Annual Meeting}} of the
  {{Association}} for {{Computational Linguistics}}}, pages 5454--5476,
  {Online}. {Association for Computational Linguistics}.

\bibitem[{Brown et~al.(2020)Brown, Mann, Ryder, Subbiah, Kaplan, Dhariwal,
  Neelakantan, Shyam, Sastry, Askell, Agarwal, Herbert-Voss, Krueger, Henighan,
  Child, Ramesh, Ziegler, Wu, Winter, Hesse, Chen, Sigler, Litwin, Gray, Chess,
  Clark, Berner, McCandlish, Radford, Sutskever, and
  Amodei}]{NEURIPS2020_1457c0d6}
Tom Brown, Benjamin Mann, Nick Ryder, Melanie Subbiah, Jared~D Kaplan, Prafulla
  Dhariwal, Arvind Neelakantan, Pranav Shyam, Girish Sastry, Amanda Askell,
  Sandhini Agarwal, Ariel Herbert-Voss, Gretchen Krueger, Tom Henighan, Rewon
  Child, Aditya Ramesh, Daniel Ziegler, Jeffrey Wu, Clemens Winter, Chris
  Hesse, Mark Chen, Eric Sigler, Mateusz Litwin, Scott Gray, Benjamin Chess,
  Jack Clark, Christopher Berner, Sam McCandlish, Alec Radford, Ilya Sutskever,
  and Dario Amodei. 2020.
\newblock \href
  {https://proceedings.neurips.cc/paper/2020/file/1457c0d6bfcb4967418bfb8ac142f64a-Paper.pdf}
  {Language models are few-shot learners}.
\newblock In \emph{Advances in Neural Information Processing Systems},
  volume~33, pages 1877--1901. Curran Associates, Inc.

\bibitem[{Carlini et~al.(2020)Carlini, Tramer, Wallace, Jagielski,
  {Herbert-Voss}, Lee, Roberts, Brown, Song, Erlingsson, Oprea, and
  Raffel}]{CarliniTramerEtAl_2020_Extracting_Training_Data_from_Large_Language_Models}
Nicholas Carlini, Florian Tramer, Eric Wallace, Matthew Jagielski, Ariel
  {Herbert-Voss}, Katherine Lee, Adam Roberts, Tom Brown, Dawn Song, Ulfar
  Erlingsson, Alina Oprea, and Colin Raffel. 2020.
\newblock \href {http://arxiv.org/abs/2012.07805} {Extracting {{Training Data}}
  from {{Large Language Models}}}.
\newblock \emph{arXiv:2012.07805 [cs]}.

\bibitem[{Chowdhery et~al.(2022)Chowdhery, Narang, Devlin, Bosma, Mishra,
  Roberts, Barham, Chung, Sutton, Gehrmann, Schuh, Shi, Tsvyashchenko, Maynez,
  Rao, Barnes, Tay, Shazeer, Prabhakaran, Reif, Du, Hutchinson, Pope, Bradbury,
  Austin, Isard, Gur-Ari, Yin, Duke, Levskaya, Ghemawat, Dev, Michalewski,
  Garcia, Misra, Robinson, Fedus, Zhou, Ippolito, Luan, Lim, Zoph, Spiridonov,
  Sepassi, Dohan, Agrawal, Omernick, Dai, Pillai, Pellat, Lewkowycz, Moreira,
  Child, Polozov, Lee, Zhou, Wang, Saeta, Diaz, Firat, Catasta, Wei,
  Meier-Hellstern, Eck, Dean, Petrov, and
  Fiedel}]{https://doi.org/10.48550/arxiv.2204.02311}
Aakanksha Chowdhery, Sharan Narang, Jacob Devlin, Maarten Bosma, Gaurav Mishra,
  Adam Roberts, Paul Barham, Hyung~Won Chung, Charles Sutton, Sebastian
  Gehrmann, Parker Schuh, Kensen Shi, Sasha Tsvyashchenko, Joshua Maynez,
  Abhishek Rao, Parker Barnes, Yi~Tay, Noam Shazeer, Vinodkumar Prabhakaran,
  Emily Reif, Nan Du, Ben Hutchinson, Reiner Pope, James Bradbury, Jacob
  Austin, Michael Isard, Guy Gur-Ari, Pengcheng Yin, Toju Duke, Anselm
  Levskaya, Sanjay Ghemawat, Sunipa Dev, Henryk Michalewski, Xavier Garcia,
  Vedant Misra, Kevin Robinson, Liam Fedus, Denny Zhou, Daphne Ippolito, David
  Luan, Hyeontaek Lim, Barret Zoph, Alexander Spiridonov, Ryan Sepassi, David
  Dohan, Shivani Agrawal, Mark Omernick, Andrew~M. Dai,
  Thanumalayan~Sankaranarayana Pillai, Marie Pellat, Aitor Lewkowycz, Erica
  Moreira, Rewon Child, Oleksandr Polozov, Katherine Lee, Zongwei Zhou, Xuezhi
  Wang, Brennan Saeta, Mark Diaz, Orhan Firat, Michele Catasta, Jason Wei,
  Kathy Meier-Hellstern, Douglas Eck, Jeff Dean, Slav Petrov, and Noah Fiedel.
  2022.
\newblock \href {https://doi.org/10.48550/ARXIV.2204.02311} {Palm: Scaling
  language modeling with pathways}.

\bibitem[{Dodge et~al.(2021)Dodge, Sap, Marasovi{\'c}, Agnew, Ilharco,
  Groeneveld, Mitchell, and
  Gardner}]{DodgeSapEtAl_2021_Documenting_Large_Webtext_Corpora_Case_Study_on_Colossal_Clean_Crawled_Corpus}
Jesse Dodge, Maarten Sap, Ana Marasovi{\'c}, William Agnew, Gabriel Ilharco,
  Dirk Groeneveld, Margaret Mitchell, and Matt Gardner. 2021.
\newblock \href {https://doi.org/10.18653/v1/2021.emnlp-main.98} {Documenting
  {{Large Webtext Corpora}}: {{A Case Study}} on the {{Colossal Clean Crawled
  Corpus}}}.
\newblock In \emph{Proceedings of the 2021 {{Conference}} on {{Empirical
  Methods}} in {{Natural Language Processing}}}, pages 1286--1305, {Online and
  Punta Cana, Dominican Republic}. {Association for Computational Linguistics}.

\bibitem[{Field et~al.(2021)Field, Blodgett, Waseem, and
  Tsvetkov}]{FieldBlodgettEtAl_2021_Survey_of_Race_Racism_and_AntiRacism_in_NLP}
Anjalie Field, Su~Lin Blodgett, Zeerak Waseem, and Yulia Tsvetkov. 2021.
\newblock \href {https://doi.org/10.18653/v1/2021.acl-long.149} {A {{Survey}}
  of {{Race}}, {{Racism}}, and {{Anti-Racism}} in {{NLP}}}.
\newblock In \emph{Proceedings of the 59th {{Annual Meeting}} of the
  {{Association}} for {{Computational Linguistics}} and the 11th
  {{International Joint Conference}} on {{Natural Language Processing}}
  ({{Volume}} 1: {{Long Papers}})}, pages 1905--1925, {Online}. {Association
  for Computational Linguistics}.

\bibitem[{Gao et~al.(2021)Gao, Biderman, Black, Golding, Hoppe, Foster, Phang,
  He, Thite, Nabeshima, Presser, and
  Leahy}]{https://doi.org/10.48550/arxiv.2101.00027}
Leo Gao, Stella Biderman, Sid Black, Laurence Golding, Travis Hoppe, Charles
  Foster, Jason Phang, Horace He, Anish Thite, Noa Nabeshima, Shawn Presser,
  and Connor Leahy. 2021.
\newblock \href {https://doi.org/10.48550/ARXIV.2101.00027} {The pile: An 800gb
  dataset of diverse text for language modeling}.

\bibitem[{Harris et~al.(2020)Harris, Millman, van~der Walt, Gommers, Virtanen,
  Cournapeau, Wieser, Taylor, Berg, Smith, Kern, Picus, Hoyer, van Kerkwijk,
  Brett, Haldane, del R{\'{i}}o, Wiebe, Peterson, G{\'{e}}rard-Marchant,
  Sheppard, Reddy, Weckesser, Abbasi, Gohlke, and Oliphant}]{harris2020array}
Charles~R. Harris, K.~Jarrod Millman, St{\'{e}}fan~J. van~der Walt, Ralf
  Gommers, Pauli Virtanen, David Cournapeau, Eric Wieser, Julian Taylor,
  Sebastian Berg, Nathaniel~J. Smith, Robert Kern, Matti Picus, Stephan Hoyer,
  Marten~H. van Kerkwijk, Matthew Brett, Allan Haldane, Jaime~Fern{\'{a}}ndez
  del R{\'{i}}o, Mark Wiebe, Pearu Peterson, Pierre G{\'{e}}rard-Marchant,
  Kevin Sheppard, Tyler Reddy, Warren Weckesser, Hameer Abbasi, Christoph
  Gohlke, and Travis~E. Oliphant. 2020.
\newblock \href {https://doi.org/10.1038/s41586-020-2649-2} {Array programming
  with {NumPy}}.
\newblock \emph{Nature}, 585(7825):357--362.

\bibitem[{Hoffmann et~al.(2022)Hoffmann, Borgeaud, Mensch, Buchatskaya, Cai,
  Rutherford, Casas, Hendricks, Welbl, Clark, Hennigan, Noland, Millican,
  Driessche, Damoc, Guy, Osindero, Simonyan, Elsen, Rae, Vinyals, and
  Sifre}]{https://doi.org/10.48550/arxiv.2203.15556}
Jordan Hoffmann, Sebastian Borgeaud, Arthur Mensch, Elena Buchatskaya, Trevor
  Cai, Eliza Rutherford, Diego de~Las Casas, Lisa~Anne Hendricks, Johannes
  Welbl, Aidan Clark, Tom Hennigan, Eric Noland, Katie Millican, George van~den
  Driessche, Bogdan Damoc, Aurelia Guy, Simon Osindero, Karen Simonyan, Erich
  Elsen, Jack~W. Rae, Oriol Vinyals, and Laurent Sifre. 2022.
\newblock \href {https://doi.org/10.48550/ARXIV.2203.15556} {Training
  compute-optimal large language models}.

\bibitem[{Huang et~al.(2022)Huang, Shao, and
  Chang}]{HuangShaoEtAl_2022_Are_Large_PreTrained_Language_Models_Leaking_Your_Personal_Information}
Jie Huang, Hanyin Shao, and Kevin Chen-Chuan Chang. 2022.
\newblock \href {https://doi.org/10.48550/arXiv.2205.12628} {Are {{Large
  Pre-Trained Language Models Leaking Your Personal Information}}?}

\bibitem[{Jernite et~al.(2022)Jernite, Nguyen, Biderman, Rogers, Masoud,
  Danchev, Tan, Luccioni, Subramani, Johnson, Dupont, Dodge, Lo, Talat, Radev,
  Gokaslan, Nikpoor, Henderson, Bommasani, and
  Mitchell}]{10.1145/3531146.3534637}
Yacine Jernite, Huu Nguyen, Stella Biderman, Anna Rogers, Maraim Masoud,
  Valentin Danchev, Samson Tan, Alexandra~Sasha Luccioni, Nishant Subramani,
  Isaac Johnson, Gerard Dupont, Jesse Dodge, Kyle Lo, Zeerak Talat, Dragomir
  Radev, Aaron Gokaslan, Somaieh Nikpoor, Peter Henderson, Rishi Bommasani, and
  Margaret Mitchell. 2022.
\newblock \href {https://doi.org/10.1145/3531146.3534637} {Data governance in
  the age of large-scale data-driven language technology}.
\newblock In \emph{2022 ACM Conference on Fairness, Accountability, and
  Transparency}, FAccT '22, page 2206–2222, New York, NY, USA. Association
  for Computing Machinery.

\bibitem[{Johnson et~al.(2019)Johnson, Douze, and
  J{\'e}gou}]{johnson2019billion-faiss}
Jeff Johnson, Matthijs Douze, and Herv{\'e} J{\'e}gou. 2019.
\newblock Billion-scale similarity search with {GPUs}.
\newblock \emph{IEEE Transactions on Big Data}, 7(3):535--547.

\bibitem[{Kaplan et~al.(2020)Kaplan, McCandlish, Henighan, Brown, Chess, Child,
  Gray, Radford, Wu, and Amodei}]{kaplan2020scaling}
Jared Kaplan, Sam McCandlish, Tom Henighan, Tom~B. Brown, Benjamin Chess, Rewon
  Child, Scott Gray, Alec Radford, Jeffrey Wu, and Dario Amodei. 2020.
\newblock \href {http://arxiv.org/abs/2001.08361} {Scaling laws for neural
  language models}.

\bibitem[{Kocetkov et~al.(2022)Kocetkov, Li, Allal, Li, Mou, Ferrandis,
  Jernite, Mitchell, Hughes, Wolf, Bahdanau, von Werra, and
  de~Vries}]{https://doi.org/10.48550/arxiv.2211.15533}
Denis Kocetkov, Raymond Li, Loubna~Ben Allal, Jia Li, Chenghao Mou,
  Carlos~Muñoz Ferrandis, Yacine Jernite, Margaret Mitchell, Sean Hughes,
  Thomas Wolf, Dzmitry Bahdanau, Leandro von Werra, and Harm de~Vries. 2022.
\newblock \href {https://doi.org/10.48550/ARXIV.2211.15533} {The stack: 3 tb of
  permissively licensed source code}.

\bibitem[{Kreutzer et~al.(2022)Kreutzer, Caswell, Wang, Wahab, {van Esch},
  {Ulzii-Orshikh}, Tapo, Subramani, Sokolov, Sikasote, Setyawan, Sarin, Samb,
  Sagot, Rivera, Rios, Papadimitriou, Osei, Suarez, Orife, Ogueji, Rubungo,
  Nguyen, M{\"u}ller, M{\"u}ller, Muhammad, Muhammad, Mnyakeni, Mirzakhalov,
  Matangira, Leong, Lawson, Kudugunta, Jernite, Jenny, Firat, Dossou, Dlamini,
  {de Silva}, {\c C}abuk~Ball{\i}, Biderman, Battisti, Baruwa, Bapna, Baljekar,
  Azime, Awokoya, Ataman, Ahia, Ahia, Agrawal, and
  Adeyemi}]{KreutzerCaswellEtAl_2022_Quality_at_Glance_Audit_of_WebCrawled_Multilingual_Datasets}
Julia Kreutzer, Isaac Caswell, Lisa Wang, Ahsan Wahab, Daan {van Esch},
  Nasanbayar {Ulzii-Orshikh}, Allahsera Tapo, Nishant Subramani, Artem Sokolov,
  Claytone Sikasote, Monang Setyawan, Supheakmungkol Sarin, Sokhar Samb,
  Beno{\^i}t Sagot, Clara Rivera, Annette Rios, Isabel Papadimitriou, Salomey
  Osei, Pedro~Ortiz Suarez, Iroro Orife, Kelechi Ogueji, Andre~Niyongabo
  Rubungo, Toan~Q. Nguyen, Mathias M{\"u}ller, Andr{\'e} M{\"u}ller,
  Shamsuddeen~Hassan Muhammad, Nanda Muhammad, Ayanda Mnyakeni, Jamshidbek
  Mirzakhalov, Tapiwanashe Matangira, Colin Leong, Nze Lawson, Sneha Kudugunta,
  Yacine Jernite, Mathias Jenny, Orhan Firat, Bonaventure F.~P. Dossou, Sakhile
  Dlamini, Nisansa {de Silva}, Sakine {\c C}abuk~Ball{\i}, Stella Biderman,
  Alessia Battisti, Ahmed Baruwa, Ankur Bapna, Pallavi Baljekar, Israel~Abebe
  Azime, Ayodele Awokoya, Duygu Ataman, Orevaoghene Ahia, Oghenefego Ahia,
  Sweta Agrawal, and Mofetoluwa Adeyemi. 2022.
\newblock \href {https://doi.org/10.1162/tacl_a_00447} {Quality at a
  {{Glance}}: {{An Audit}} of {{Web-Crawled Multilingual Datasets}}}.
\newblock \emph{Transactions of the Association for Computational Linguistics},
  10:50--72.

\bibitem[{Lauren{\c{c}}on et~al.(2022)Lauren{\c{c}}on, Saulnier, Wang, Akiki,
  del Moral, Scao, Werra, Mou, Ponferrada, Nguyen, Frohberg, {\v{S}}a{\v{s}}ko,
  Lhoest, McMillan-Major, Dupont, Biderman, Rogers, allal, Toni, Pistilli,
  Nguyen, Nikpoor, Masoud, Colombo, de~la Rosa, Villegas, Thrush, Longpre,
  Nagel, Weber, Mu{\~n}oz, Zhu, Strien, Alyafeai, Almubarak, Chien,
  Gonzalez-Dios, Soroa, Lo, Dey, Suarez, Gokaslan, Bose, Adelani, Phan, Tran,
  Yu, Pai, Chim, Lepercq, Ilic, Mitchell, Luccioni, and
  Jernite}]{laurencon2022the}
Hugo Lauren{\c{c}}on, Lucile Saulnier, Thomas Wang, Christopher Akiki,
  Albert~Villanova del Moral, Teven~Le Scao, Leandro~Von Werra, Chenghao Mou,
  Eduardo~Gonz{\'a}lez Ponferrada, Huu Nguyen, J{\"o}rg Frohberg, Mario
  {\v{S}}a{\v{s}}ko, Quentin Lhoest, Angelina McMillan-Major, G{\'e}rard
  Dupont, Stella Biderman, Anna Rogers, Loubna~Ben allal, Francesco~De Toni,
  Giada Pistilli, Olivier Nguyen, Somaieh Nikpoor, Maraim Masoud, Pierre
  Colombo, Javier de~la Rosa, Paulo Villegas, Tristan Thrush, Shayne Longpre,
  Sebastian Nagel, Leon Weber, Manuel~Romero Mu{\~n}oz, Jian Zhu, Daniel~Van
  Strien, Zaid Alyafeai, Khalid Almubarak, Vu~Minh Chien, Itziar Gonzalez-Dios,
  Aitor Soroa, Kyle Lo, Manan Dey, Pedro~Ortiz Suarez, Aaron Gokaslan, Shamik
  Bose, David~Ifeoluwa Adelani, Long Phan, Hieu Tran, Ian Yu, Suhas Pai, Jenny
  Chim, Violette Lepercq, Suzana Ilic, Margaret Mitchell, Sasha Luccioni, and
  Yacine Jernite. 2022.
\newblock \href {https://openreview.net/forum?id=UoEw6KigkUn} {The bigscience
  {ROOTS} corpus: A 1.6{TB} composite multilingual dataset}.
\newblock In \emph{Thirty-sixth Conference on Neural Information Processing
  Systems Datasets and Benchmarks Track}.

\bibitem[{Le~Scao et~al.(2022)Le~Scao, Fan, Akiki, Pavlick, Ili{\'c}, Hesslow,
  Castagn{\'e}, Luccioni, Yvon, Gall{\'e}, Tow, Rush, Biderman, Webson,
  Ammanamanchi, Wang, Sagot, Muennighoff, {del Moral}, Ruwase, Bawden, Bekman,
  {McMillan-Major}, Beltagy, Nguyen, Saulnier, Tan, Suarez, Sanh, Lauren{\c
  c}on, Jernite, Launay, Mitchell, Raffel, Gokaslan, Simhi, Soroa, Aji,
  Alfassy, Rogers, Nitzav, Xu, Mou, Emezue, Klamm, Leong, {van Strien},
  Adelani, Radev, Ponferrada, Levkovizh, Kim, Natan, De~Toni, Dupont,
  Kruszewski, Pistilli, Elsahar, Benyamina, Tran, Yu, Abdulmumin, Johnson,
  {Gonzalez-Dios}, {de la Rosa}, Chim, Dodge, Zhu, Chang, Frohberg, Tobing,
  Bhattacharjee, Almubarak, Chen, Lo, Von~Werra, Weber, Phan, {allal}, Tanguy,
  Dey, Mu{\~n}oz, Masoud, Grandury, {\v S}a{\v s}ko, Huang, Coavoux, Singh,
  Jiang, Vu, Jauhar, Ghaleb, Subramani, Kassner, Khamis, Nguyen, Espejel, {de
  Gibert}, Villegas, Henderson, Colombo, Amuok, Lhoest, Harliman, Bommasani,
  L{\'o}pez, Ribeiro, Osei, Pyysalo, Nagel, Bose, Muhammad, Sharma, Longpre,
  Nikpoor, Silberberg, Pai, Zink, Torrent, Schick, Thrush, Danchev, Nikoulina,
  Laippala, Lepercq, Prabhu, Alyafeai, Talat, Raja, Heinzerling, Si, Ta{\c
  s}ar, Salesky, Mielke, Lee, Sharma, Santilli, Chaffin, Stiegler, Datta,
  Szczechla, Chhablani, Wang, Pandey, Strobelt, Fries, Rozen, Gao, Sutawika,
  Bari, {Al-shaibani}, Manica, Nayak, Teehan, Albanie, Shen, {Ben-David}, Bach,
  Kim, Bers, Fevry, Neeraj, Thakker, Raunak, Tang, Yong, Sun, Brody, Uri,
  Tojarieh, Roberts, Chung, Tae, Phang, Press, Li, Narayanan, Bourfoune,
  Casper, Rasley, Ryabinin, Mishra, Zhang, Shoeybi, Peyrounette, Patry, Tazi,
  Sanseviero, {von Platen}, Cornette, Lavall{\'e}e, Lacroix, Rajbhandari,
  Gandhi, Smith, Requena, Patil, Dettmers, Baruwa, Singh, Cheveleva, Ligozat,
  Subramonian, N{\'e}v{\'e}ol, Lovering, Garrette, Tunuguntla, Reiter,
  Taktasheva, Voloshina, Bogdanov, Winata, Schoelkopf, Kalo, Novikova, Forde,
  Clive, Kasai, Kawamura, Hazan, Carpuat, Clinciu, Kim, Cheng, Serikov,
  Antverg, {van der Wal}, Zhang, Zhang, Gehrmann, Mirkin, Pais, Shavrina,
  Scialom, Yun, Limisiewicz, Rieser, Protasov, Mikhailov, Pruksachatkun,
  Belinkov, Bamberger, Kasner, Rueda, Pestana, Feizpour, Khan, Faranak, Santos,
  Hevia, Unldreaj, Aghagol, Abdollahi, Tammour, HajiHosseini, Behroozi,
  Ajibade, Saxena, Ferrandis, Contractor, Lansky, David, Kiela, Nguyen, Tan,
  Baylor, Ozoani, Mirza, Ononiwu, Rezanejad, Jones, Bhattacharya, Solaiman,
  Sedenko, Nejadgholi, Passmore, Seltzer, Sanz, Dutra, Samagaio, Elbadri,
  Mieskes, Gerchick, Akinlolu, McKenna, Qiu, Ghauri, Burynok, Abrar, Rajani,
  Elkott, Fahmy, Samuel, An, Kromann, Hao, Alizadeh, Shubber, Wang, Roy,
  Viguier, Le, Oyebade, Le, Yang, Nguyen, Kashyap, Palasciano, Callahan,
  Shukla, {Miranda-Escalada}, Singh, Beilharz, Wang, Brito, Zhou, Jain, Xu,
  Fourrier, Peri{\~n}{\'a}n, Molano, Yu, Manjavacas, Barth, Fuhrimann, Altay,
  Bayrak, Burns, Vrabec, Bello, Dash, Kang, Giorgi, Golde, Posada, Sivaraman,
  Bulchandani, Liu, Shinzato, {de Bykhovetz}, Takeuchi, P{\`a}mies, Castillo,
  Nezhurina, S{\"a}nger, Samwald, Cullan, Weinberg, De~Wolf, Mihaljcic, Liu,
  Freidank, Kang, Seelam, Dahlberg, Broad, Muellner, Fung, Haller,
  Chandrasekhar, Eisenberg, Martin, Canalli, Su, Su, Cahyawijaya, Garda,
  Deshmukh, Mishra, Kiblawi, Ott, {Sang-aroonsiri}, Kumar, Schweter, Bharati,
  Laud, Gigant, Kainuma, Kusa, Labrak, Bajaj, Venkatraman, Xu, Xu, Xu, Tan,
  Xie, Ye, Bras, Belkada, and Wolf}]{bloom}
Teven Le~Scao, Angela Fan, Christopher Akiki, Ellie Pavlick, Suzana Ili{\'c},
  Daniel Hesslow, Roman Castagn{\'e}, Alexandra~Sasha Luccioni, Fran{\c c}ois
  Yvon, Matthias Gall{\'e}, Jonathan Tow, Alexander~M. Rush, Stella Biderman,
  Albert Webson, Pawan~Sasanka Ammanamanchi, Thomas Wang, Beno{\^i}t Sagot,
  Niklas Muennighoff, Albert~Villanova {del Moral}, Olatunji Ruwase, Rachel
  Bawden, Stas Bekman, Angelina {McMillan-Major}, Iz~Beltagy, Huu Nguyen,
  Lucile Saulnier, Samson Tan, Pedro~Ortiz Suarez, Victor Sanh, Hugo Lauren{\c
  c}on, Yacine Jernite, Julien Launay, Margaret Mitchell, Colin Raffel, Aaron
  Gokaslan, Adi Simhi, Aitor Soroa, Alham~Fikri Aji, Amit Alfassy, Anna Rogers,
  Ariel~Kreisberg Nitzav, Canwen Xu, Chenghao Mou, Chris Emezue, Christopher
  Klamm, Colin Leong, Daniel {van Strien}, David~Ifeoluwa Adelani, Dragomir
  Radev, Eduardo~Gonz{\'a}lez Ponferrada, Efrat Levkovizh, Ethan Kim, Eyal~Bar
  Natan, Francesco De~Toni, G{\'e}rard Dupont, Germ{\'a}n Kruszewski, Giada
  Pistilli, Hady Elsahar, Hamza Benyamina, Hieu Tran, Ian Yu, Idris Abdulmumin,
  Isaac Johnson, Itziar {Gonzalez-Dios}, Javier {de la Rosa}, Jenny Chim, Jesse
  Dodge, Jian Zhu, Jonathan Chang, J{\"o}rg Frohberg, Joseph Tobing, Joydeep
  Bhattacharjee, Khalid Almubarak, Kimbo Chen, Kyle Lo, Leandro Von~Werra, Leon
  Weber, Long Phan, Loubna~Ben {allal}, Ludovic Tanguy, Manan Dey,
  Manuel~Romero Mu{\~n}oz, Maraim Masoud, Mar{\'i}a Grandury, Mario {\v S}a{\v
  s}ko, Max Huang, Maximin Coavoux, Mayank Singh, Mike Tian-Jian Jiang,
  Minh~Chien Vu, Mohammad~A. Jauhar, Mustafa Ghaleb, Nishant Subramani, Nora
  Kassner, Nurulaqilla Khamis, Olivier Nguyen, Omar Espejel, Ona {de Gibert},
  Paulo Villegas, Peter Henderson, Pierre Colombo, Priscilla Amuok, Quentin
  Lhoest, Rheza Harliman, Rishi Bommasani, Roberto~Luis L{\'o}pez, Rui Ribeiro,
  Salomey Osei, Sampo Pyysalo, Sebastian Nagel, Shamik Bose, Shamsuddeen~Hassan
  Muhammad, Shanya Sharma, Shayne Longpre, Somaieh Nikpoor, Stanislav
  Silberberg, Suhas Pai, Sydney Zink, Tiago~Timponi Torrent, Timo Schick,
  Tristan Thrush, Valentin Danchev, Vassilina Nikoulina, Veronika Laippala,
  Violette Lepercq, Vrinda Prabhu, Zaid Alyafeai, Zeerak Talat, Arun Raja,
  Benjamin Heinzerling, Chenglei Si, Davut~Emre Ta{\c s}ar, Elizabeth Salesky,
  Sabrina~J. Mielke, Wilson~Y. Lee, Abheesht Sharma, Andrea Santilli, Antoine
  Chaffin, Arnaud Stiegler, Debajyoti Datta, Eliza Szczechla, Gunjan Chhablani,
  Han Wang, Harshit Pandey, Hendrik Strobelt, Jason~Alan Fries, Jos Rozen, Leo
  Gao, Lintang Sutawika, M.~Saiful Bari, Maged~S. {Al-shaibani}, Matteo Manica,
  Nihal Nayak, Ryan Teehan, Samuel Albanie, Sheng Shen, Srulik {Ben-David},
  Stephen~H. Bach, Taewoon Kim, Tali Bers, Thibault Fevry, Trishala Neeraj,
  Urmish Thakker, Vikas Raunak, Xiangru Tang, Zheng-Xin Yong, Zhiqing Sun,
  Shaked Brody, Yallow Uri, Hadar Tojarieh, Adam Roberts, Hyung~Won Chung,
  Jaesung Tae, Jason Phang, Ofir Press, Conglong Li, Deepak Narayanan, Hatim
  Bourfoune, Jared Casper, Jeff Rasley, Max Ryabinin, Mayank Mishra, Minjia
  Zhang, Mohammad Shoeybi, Myriam Peyrounette, Nicolas Patry, Nouamane Tazi,
  Omar Sanseviero, Patrick {von Platen}, Pierre Cornette, Pierre~Fran{\c c}ois
  Lavall{\'e}e, R{\'e}mi Lacroix, Samyam Rajbhandari, Sanchit Gandhi, Shaden
  Smith, St{\'e}phane Requena, Suraj Patil, Tim Dettmers, Ahmed Baruwa,
  Amanpreet Singh, Anastasia Cheveleva, Anne-Laure Ligozat, Arjun Subramonian,
  Aur{\'e}lie N{\'e}v{\'e}ol, Charles Lovering, Dan Garrette, Deepak
  Tunuguntla, Ehud Reiter, Ekaterina Taktasheva, Ekaterina Voloshina, Eli
  Bogdanov, Genta~Indra Winata, Hailey Schoelkopf, Jan-Christoph Kalo,
  Jekaterina Novikova, Jessica~Zosa Forde, Jordan Clive, Jungo Kasai, Ken
  Kawamura, Liam Hazan, Marine Carpuat, Miruna Clinciu, Najoung Kim, Newton
  Cheng, Oleg Serikov, Omer Antverg, Oskar {van der Wal}, Rui Zhang, Ruochen
  Zhang, Sebastian Gehrmann, Shachar Mirkin, Shani Pais, Tatiana Shavrina,
  Thomas Scialom, Tian Yun, Tomasz Limisiewicz, Verena Rieser, Vitaly Protasov,
  Vladislav Mikhailov, Yada Pruksachatkun, Yonatan Belinkov, Zachary Bamberger,
  Zden{\v e}k Kasner, Alice Rueda, Amanda Pestana, Amir Feizpour, Ammar Khan,
  Amy Faranak, Ana Santos, Anthony Hevia, Antigona Unldreaj, Arash Aghagol,
  Arezoo Abdollahi, Aycha Tammour, Azadeh HajiHosseini, Bahareh Behroozi,
  Benjamin Ajibade, Bharat Saxena, Carlos~Mu{\~n}oz Ferrandis, Danish
  Contractor, David Lansky, Davis David, Douwe Kiela, Duong~A. Nguyen, Edward
  Tan, Emi Baylor, Ezinwanne Ozoani, Fatima Mirza, Frankline Ononiwu, Habib
  Rezanejad, Hessie Jones, Indrani Bhattacharya, Irene Solaiman, Irina Sedenko,
  Isar Nejadgholi, Jesse Passmore, Josh Seltzer, Julio~Bonis Sanz, Livia Dutra,
  Mairon Samagaio, Maraim Elbadri, Margot Mieskes, Marissa Gerchick, Martha
  Akinlolu, Michael McKenna, Mike Qiu, Muhammed Ghauri, Mykola Burynok, Nafis
  Abrar, Nazneen Rajani, Nour Elkott, Nour Fahmy, Olanrewaju Samuel, Ran An,
  Rasmus Kromann, Ryan Hao, Samira Alizadeh, Sarmad Shubber, Silas Wang, Sourav
  Roy, Sylvain Viguier, Thanh Le, Tobi Oyebade, Trieu Le, Yoyo Yang, Zach
  Nguyen, Abhinav~Ramesh Kashyap, Alfredo Palasciano, Alison Callahan, Anima
  Shukla, Antonio {Miranda-Escalada}, Ayush Singh, Benjamin Beilharz, Bo~Wang,
  Caio Brito, Chenxi Zhou, Chirag Jain, Chuxin Xu, Cl{\'e}mentine Fourrier,
  Daniel~Le{\'o}n Peri{\~n}{\'a}n, Daniel Molano, Dian Yu, Enrique Manjavacas,
  Fabio Barth, Florian Fuhrimann, Gabriel Altay, Giyaseddin Bayrak, Gully
  Burns, Helena~U. Vrabec, Imane Bello, Ishani Dash, Jihyun Kang, John Giorgi,
  Jonas Golde, Jose~David Posada, Karthik~Rangasai Sivaraman, Lokesh
  Bulchandani, Lu~Liu, Luisa Shinzato, Madeleine~Hahn {de Bykhovetz}, Maiko
  Takeuchi, Marc P{\`a}mies, Maria~A. Castillo, Marianna Nezhurina, Mario
  S{\"a}nger, Matthias Samwald, Michael Cullan, Michael Weinberg, Michiel
  De~Wolf, Mina Mihaljcic, Minna Liu, Moritz Freidank, Myungsun Kang, Natasha
  Seelam, Nathan Dahlberg, Nicholas~Michio Broad, Nikolaus Muellner, Pascale
  Fung, Patrick Haller, Ramya Chandrasekhar, Renata Eisenberg, Robert Martin,
  Rodrigo Canalli, Rosaline Su, Ruisi Su, Samuel Cahyawijaya, Samuele Garda,
  Shlok~S. Deshmukh, Shubhanshu Mishra, Sid Kiblawi, Simon Ott, Sinee
  {Sang-aroonsiri}, Srishti Kumar, Stefan Schweter, Sushil Bharati, Tanmay
  Laud, Th{\'e}o Gigant, Tomoya Kainuma, Wojciech Kusa, Yanis Labrak,
  Yash~Shailesh Bajaj, Yash Venkatraman, Yifan Xu, Yingxin Xu, Yu~Xu, Zhe Tan,
  Zhongli Xie, Zifan Ye, Mathilde Bras, Younes Belkada, and Thomas Wolf. 2022.
\newblock \href {https://doi.org/10.48550/arXiv.2211.05100} {{{BLOOM}}: {{A
  176B-Parameter Open-Access Multilingual Language Model}}}.
\newblock In \emph{Thirty-Sixth {{Conference}} on {{Neural Information
  Processing Systems}}}, {New Orleans, Louisiana}. arXiv.

\bibitem[{Lee et~al.(2022)Lee, Ippolito, Nystrom, Zhang, Eck, Callison-Burch,
  and Carlini}]{lee-etal-2022-deduplicating}
Katherine Lee, Daphne Ippolito, Andrew Nystrom, Chiyuan Zhang, Douglas Eck,
  Chris Callison-Burch, and Nicholas Carlini. 2022.
\newblock \href {https://doi.org/10.18653/v1/2022.acl-long.577} {Deduplicating
  training data makes language models better}.
\newblock In \emph{Proceedings of the 60th Annual Meeting of the Association
  for Computational Linguistics (Volume 1: Long Papers)}, pages 8424--8445,
  Dublin, Ireland. Association for Computational Linguistics.

\bibitem[{Lhoest et~al.(2021)Lhoest, Villanova~del Moral, Jernite, Thakur, von
  Platen, Patil, Chaumond, Drame, Plu, Tunstall, Davison, {\v{S}}a{\v{s}}ko,
  Chhablani, Malik, Brandeis, Le~Scao, Sanh, Xu, Patry, McMillan-Major, Schmid,
  Gugger, Delangue, Matussi{\`e}re, Debut, Bekman, Cistac, Goehringer, Mustar,
  Lagunas, Rush, and Wolf}]{lhoest-etal-2021-datasets}
Quentin Lhoest, Albert Villanova~del Moral, Yacine Jernite, Abhishek Thakur,
  Patrick von Platen, Suraj Patil, Julien Chaumond, Mariama Drame, Julien Plu,
  Lewis Tunstall, Joe Davison, Mario {\v{S}}a{\v{s}}ko, Gunjan Chhablani,
  Bhavitvya Malik, Simon Brandeis, Teven Le~Scao, Victor Sanh, Canwen Xu,
  Nicolas Patry, Angelina McMillan-Major, Philipp Schmid, Sylvain Gugger,
  Cl{\'e}ment Delangue, Th{\'e}o Matussi{\`e}re, Lysandre Debut, Stas Bekman,
  Pierric Cistac, Thibault Goehringer, Victor Mustar, Fran{\c{c}}ois Lagunas,
  Alexander Rush, and Thomas Wolf. 2021.
\newblock \href {https://doi.org/10.18653/v1/2021.emnlp-demo.21} {Datasets: A
  community library for natural language processing}.
\newblock In \emph{Proceedings of the 2021 Conference on Empirical Methods in
  Natural Language Processing: System Demonstrations}, pages 175--184, Online
  and Punta Cana, Dominican Republic. Association for Computational
  Linguistics.

\bibitem[{Lieber et~al.(2021)Lieber, Sharir, Lenz, and Shoham}]{J1WhitePaper}
Opher Lieber, Or~Sharir, Barak Lenz, and Yoav Shoham. 2021.
\newblock Jurassic-1: Technical details and evaluation.
\newblock Technical report, AI21 Labs.

\bibitem[{Lin et~al.(2021)Lin, Ma, Lin, Yang, Pradeep, and
  Nogueira}]{Lin_etal_SIGIR2021_Pyserini}
Jimmy Lin, Xueguang Ma, Sheng-Chieh Lin, Jheng-Hong Yang, Ronak Pradeep, and
  Rodrigo Nogueira. 2021.
\newblock {Pyserini}: A {Python} toolkit for reproducible information retrieval
  research with sparse and dense representations.
\newblock In \emph{Proceedings of the 44th Annual International ACM SIGIR
  Conference on Research and Development in Information Retrieval (SIGIR
  2021)}, pages 2356--2362.

\bibitem[{Luccioni and Viviano(2021)}]{luccioni-viviano-2021-whats}
Alexandra Luccioni and Joseph Viviano. 2021.
\newblock \href {https://doi.org/10.18653/v1/2021.acl-short.24} {What{'}s in
  the box? an analysis of undesirable content in the {C}ommon {C}rawl corpus}.
\newblock In \emph{Proceedings of the 59th Annual Meeting of the Association
  for Computational Linguistics and the 11th International Joint Conference on
  Natural Language Processing (Volume 2: Short Papers)}, pages 182--189,
  Online. Association for Computational Linguistics.

\bibitem[{Mielke et~al.(2021)Mielke, Alyafeai, Salesky, Raffel, Dey, Gall{\'e},
  Raja, Si, Lee, Sagot, and Tan}]{Mielke2021BetweenWA}
Sabrina~J. Mielke, Zaid Alyafeai, Elizabeth Salesky, Colin Raffel, Manan Dey,
  Matthias Gall{\'e}, Arun Raja, Chenglei Si, Wilson~Y. Lee, Beno{\^i}t Sagot,
  and Samson Tan. 2021.
\newblock Between words and characters: A brief history of open-vocabulary
  modeling and tokenization in nlp.
\newblock \emph{ArXiv}, abs/2112.10508.

\bibitem[{MOI et~al.(2022)MOI, Patry, Cistac, Pete, Morgan, Pütz, Mishig,
  Johansen, Wolf, Gugger, Clement, Chaumond, Debut, Garillot, Georges, dctelus,
  Louis, MarcusGrass, Peyash, 0xflotus, deLevie, Mamaev, Arthur, Cameron,
  Clement, Moges, Hewitt, Zolotukhin, and Thomas}]{anthony_moi_2022_tokenizers}
Anthony MOI, Nicolas Patry, Pierric Cistac, Pete, Funtowicz Morgan, Sebastian
  Pütz, Mishig, Bjarte Johansen, Thomas Wolf, Sylvain Gugger, Clement, Julien
  Chaumond, Lysandre Debut, François Garillot, Luc Georges, dctelus, JC~Louis,
  MarcusGrass, Taufiquzzaman Peyash, 0xflotus, Alan deLevie, Alexander Mamaev,
  Arthur, Cameron, Colin Clement, Dagmawi Moges, David Hewitt, Denis
  Zolotukhin, and Geoffrey Thomas. 2022.
\newblock \href {https://doi.org/10.5281/zenodo.7298413}
  {huggingface/tokenizers: Rust 0.13.2}.

\bibitem[{Niezni et~al.(2022)Niezni, Taub-Tabib, Harris, Sason-Bauer, Amrusi,
  Azagury, Avrashami, Launer-Wachs, Borchardt, Kusold, Tiktinsky, Hope,
  Goldberg, and Shamay}]{cancer-nlp-no-code}
Danna Niezni, Hillel Taub-Tabib, Yuval Harris, Hagit Sason-Bauer, Yakir Amrusi,
  Dana Azagury, Maytal Avrashami, Shaked Launer-Wachs, Jon Borchardt, M~Kusold,
  Aryeh Tiktinsky, Tom Hope, Yoav Goldberg, and Yosi Shamay. 2022.
\newblock \href {https://doi.org/10.1101/2022.05.03.490286} {Extending the
  boundaries of cancer therapeutic complexity with literature data mining}.
\newblock \emph{bioRxiv}.

\bibitem[{Ogundepo et~al.(2022)Ogundepo, Zhang, and
  Lin}]{https://doi.org/10.48550/arxiv.2210.05481}
Odunayo Ogundepo, Xinyu Zhang, and Jimmy Lin. 2022.
\newblock \href {https://doi.org/10.48550/ARXIV.2210.05481} {Better than
  whitespace: Information retrieval for languages without custom tokenizers}.

\bibitem[{pandas~development team(2020)}]{reback2020pandas}
The pandas~development team. 2020.
\newblock \href {https://doi.org/10.5281/zenodo.3509134} {pandas-dev/pandas:
  Pandas}.

\bibitem[{Rae et~al.(2021)Rae, Borgeaud, Cai, Millican, Hoffmann, Song,
  Aslanides, Henderson, Ring, Young, Rutherford, Hennigan, Menick, Cassirer,
  Powell, van~den Driessche, Hendricks, Rauh, Huang, Glaese, Welbl, Dathathri,
  Huang, Uesato, Mellor, Higgins, Creswell, McAleese, Wu, Elsen, Jayakumar,
  Buchatskaya, Budden, Sutherland, Simonyan, Paganini, Sifre, Martens, Li,
  Kuncoro, Nematzadeh, Gribovskaya, Donato, Lazaridou, Mensch, Lespiau,
  Tsimpoukelli, Grigorev, Fritz, Sottiaux, Pajarskas, Pohlen, Gong, Toyama,
  de~Masson~d'Autume, Li, Terzi, Mikulik, Babuschkin, Clark, de~Las~Casas, Guy,
  Jones, Bradbury, Johnson, Hechtman, Weidinger, Gabriel, Isaac, Lockhart,
  Osindero, Rimell, Dyer, Vinyals, Ayoub, Stanway, Bennett, Hassabis,
  Kavukcuoglu, and Irving}]{Rae2021ScalingLM}
Jack~W. Rae, Sebastian Borgeaud, Trevor Cai, Katie Millican, Jordan Hoffmann,
  Francis Song, John Aslanides, Sarah Henderson, Roman Ring, Susannah Young,
  Eliza Rutherford, Tom Hennigan, Jacob Menick, Albin Cassirer, Richard Powell,
  George van~den Driessche, Lisa~Anne Hendricks, Maribeth Rauh, Po-Sen Huang,
  Amelia Glaese, Johannes Welbl, Sumanth Dathathri, Saffron Huang, Jonathan
  Uesato, John F.~J. Mellor, Irina Higgins, Antonia Creswell, Nathan McAleese,
  Amy Wu, Erich Elsen, Siddhant~M. Jayakumar, Elena Buchatskaya, David Budden,
  Esme Sutherland, Karen Simonyan, Michela Paganini, L.~Sifre, Lena Martens,
  Xiang~Lorraine Li, Adhiguna Kuncoro, Aida Nematzadeh, Elena Gribovskaya,
  Domenic Donato, Angeliki Lazaridou, Arthur Mensch, Jean-Baptiste Lespiau,
  Maria Tsimpoukelli, N.~K. Grigorev, Doug Fritz, Thibault Sottiaux, Mantas
  Pajarskas, Tobias Pohlen, Zhitao Gong, Daniel Toyama, Cyprien
  de~Masson~d'Autume, Yujia Li, Tayfun Terzi, Vladimir Mikulik, Igor
  Babuschkin, Aidan Clark, Diego de~Las~Casas, Aurelia Guy, Chris Jones, James
  Bradbury, Matthew~G. Johnson, Blake~A. Hechtman, Laura Weidinger, Iason
  Gabriel, William~S. Isaac, Edward Lockhart, Simon Osindero, Laura Rimell,
  Chris Dyer, Oriol Vinyals, Kareem~W. Ayoub, Jeff Stanway, L.~L. Bennett,
  Demis Hassabis, Koray Kavukcuoglu, and Geoffrey Irving. 2021.
\newblock Scaling language models: Methods, analysis \& insights from training
  gopher.
\newblock \emph{ArXiv}, abs/2112.11446.

\bibitem[{Raffel et~al.(2020)Raffel, Shazeer, Roberts, Lee, Narang, Matena,
  Zhou, Li, and Liu}]{10.5555/3455716.3455856}
Colin Raffel, Noam Shazeer, Adam Roberts, Katherine Lee, Sharan Narang, Michael
  Matena, Yanqi Zhou, Wei Li, and Peter~J. Liu. 2020.
\newblock Exploring the limits of transfer learning with a unified text-to-text
  transformer.
\newblock \emph{J. Mach. Learn. Res.}, 21(1).

\bibitem[{Ramesh et~al.(2022)Ramesh, Dhariwal, Nichol, Chu, and
  Chen}]{https://doi.org/10.48550/arxiv.2204.06125}
Aditya Ramesh, Prafulla Dhariwal, Alex Nichol, Casey Chu, and Mark Chen. 2022.
\newblock \href {https://doi.org/10.48550/ARXIV.2204.06125} {Hierarchical
  text-conditional image generation with clip latents}.

\bibitem[{Rombach et~al.(2021)Rombach, Blattmann, Lorenz, Esser, and
  Ommer}]{rombach2021highresolution}
Robin Rombach, Andreas Blattmann, Dominik Lorenz, Patrick Esser, and Björn
  Ommer. 2021.
\newblock \href {http://arxiv.org/abs/2112.10752} {High-resolution image
  synthesis with latent diffusion models}.

\bibitem[{Schuhmann et~al.(2022)Schuhmann, Beaumont, Vencu, Gordon, Wightman,
  Cherti, Coombes, Katta, Mullis, Wortsman, Schramowski, Kundurthy, Crowson,
  Schmidt, Kaczmarczyk, and Jitsev}]{https://doi.org/10.48550/arxiv.2210.08402}
Christoph Schuhmann, Romain Beaumont, Richard Vencu, Cade Gordon, Ross
  Wightman, Mehdi Cherti, Theo Coombes, Aarush Katta, Clayton Mullis, Mitchell
  Wortsman, Patrick Schramowski, Srivatsa Kundurthy, Katherine Crowson, Ludwig
  Schmidt, Robert Kaczmarczyk, and Jenia Jitsev. 2022.
\newblock \href {https://doi.org/10.48550/ARXIV.2210.08402} {Laion-5b: An open
  large-scale dataset for training next generation image-text models}.

\bibitem[{Smith et~al.(2015)Smith, Cordell, and Mullen}]{10.1093/alh/ajv029}
David~A. Smith, Ryan Cordell, and Abby Mullen. 2015.
\newblock \href {https://doi.org/10.1093/alh/ajv029} {{Computational Methods
  for Uncovering Reprinted Texts in Antebellum Newspapers}}.
\newblock \emph{American Literary History}, 27(3):E1--E15.

\bibitem[{Smith et~al.(2022)Smith, Patwary, Norick, LeGresley, Rajbhandari,
  Casper, Liu, Prabhumoye, Zerveas, Korthikanti, Zhang, Child, Aminabadi,
  Bernauer, Song, Shoeybi, He, Houston, Tiwary, and
  Catanzaro}]{https://doi.org/10.48550/arxiv.2201.11990}
Shaden Smith, Mostofa Patwary, Brandon Norick, Patrick LeGresley, Samyam
  Rajbhandari, Jared Casper, Zhun Liu, Shrimai Prabhumoye, George Zerveas,
  Vijay Korthikanti, Elton Zhang, Rewon Child, Reza~Yazdani Aminabadi, Julie
  Bernauer, Xia Song, Mohammad Shoeybi, Yuxiong He, Michael Houston, Saurabh
  Tiwary, and Bryan Catanzaro. 2022.
\newblock \href {https://doi.org/10.48550/ARXIV.2201.11990} {Using deepspeed
  and megatron to train megatron-turing nlg 530b, a large-scale generative
  language model}.

\bibitem[{Stanczak and
  Augenstein(2021)}]{StanczakAugenstein_2021_Survey_on_Gender_Bias_in_Natural_Language_Processing}
Karolina Stanczak and Isabelle Augenstein. 2021.
\newblock \href {https://doi.org/10.48550/arXiv.2112.14168} {A {{Survey}} on
  {{Gender Bias}} in {{Natural Language Processing}}}.

\bibitem[{Tang(2021)}]{WuDao}
Jie Tang. 2021.
\newblock {WuDao}: Pretrain the world.
\newblock Keynote adress at the European Conference on Machine Learning and
  Principles and Practice of Knowledge Discovery in Databases.

\bibitem[{Thoppilan et~al.(2022)Thoppilan, De~Freitas, Hall, Shazeer,
  Kulshreshtha, Cheng, Jin, Bos, Baker, Du, Li, Lee, Zheng, Ghafouri, Menegali,
  Huang, Krikun, Lepikhin, Qin, Chen, Xu, Chen, Roberts, Bosma, Zhao, Zhou,
  Chang, Krivokon, Rusch, Pickett, Srinivasan, Man, Meier-Hellstern, Morris,
  Doshi, Santos, Duke, Soraker, Zevenbergen, Prabhakaran, Diaz, Hutchinson,
  Olson, Molina, Hoffman-John, Lee, Aroyo, Rajakumar, Butryna, Lamm, Kuzmina,
  Fenton, Cohen, Bernstein, Kurzweil, Aguera-Arcas, Cui, Croak, Chi, and
  Le}]{https://doi.org/10.48550/arxiv.2201.08239}
Romal Thoppilan, Daniel De~Freitas, Jamie Hall, Noam Shazeer, Apoorv
  Kulshreshtha, Heng-Tze Cheng, Alicia Jin, Taylor Bos, Leslie Baker, Yu~Du,
  YaGuang Li, Hongrae Lee, Huaixiu~Steven Zheng, Amin Ghafouri, Marcelo
  Menegali, Yanping Huang, Maxim Krikun, Dmitry Lepikhin, James Qin, Dehao
  Chen, Yuanzhong Xu, Zhifeng Chen, Adam Roberts, Maarten Bosma, Vincent Zhao,
  Yanqi Zhou, Chung-Ching Chang, Igor Krivokon, Will Rusch, Marc Pickett,
  Pranesh Srinivasan, Laichee Man, Kathleen Meier-Hellstern, Meredith~Ringel
  Morris, Tulsee Doshi, Renelito~Delos Santos, Toju Duke, Johnny Soraker, Ben
  Zevenbergen, Vinodkumar Prabhakaran, Mark Diaz, Ben Hutchinson, Kristen
  Olson, Alejandra Molina, Erin Hoffman-John, Josh Lee, Lora Aroyo, Ravi
  Rajakumar, Alena Butryna, Matthew Lamm, Viktoriya Kuzmina, Joe Fenton, Aaron
  Cohen, Rachel Bernstein, Ray Kurzweil, Blaise Aguera-Arcas, Claire Cui,
  Marian Croak, Ed~Chi, and Quoc Le. 2022.
\newblock \href {https://doi.org/10.48550/ARXIV.2201.08239} {Lamda: Language
  models for dialog applications}.

\bibitem[{Touvron et~al.(2023)Touvron, Lavril, Izacard, Martinet, Lachaux,
  Lacroix, Rozière, Goyal, Hambro, Azhar, Rodriguez, Joulin, Grave, and
  Lample}]{touvron2023llama}
Hugo Touvron, Thibaut Lavril, Gautier Izacard, Xavier Martinet, Marie-Anne
  Lachaux, Timothée Lacroix, Baptiste Rozière, Naman Goyal, Eric Hambro,
  Faisal Azhar, Aurelien Rodriguez, Armand Joulin, Edouard Grave, and Guillaume
  Lample. 2023.
\newblock \href {http://arxiv.org/abs/2302.13971} {Llama: Open and efficient
  foundation language models}.

\bibitem[{Virtanen et~al.(2020)Virtanen, Gommers, Oliphant, Haberland, Reddy,
  Cournapeau, Burovski, Peterson, Weckesser, Bright, {van der Walt}, Brett,
  Wilson, Millman, Mayorov, Nelson, Jones, Kern, Larson, Carey, Polat, Feng,
  Moore, {VanderPlas}, Laxalde, Perktold, Cimrman, Henriksen, Quintero, Harris,
  Archibald, Ribeiro, Pedregosa, {van Mulbregt}, and {SciPy 1.0
  Contributors}}]{2020SciPy-NMeth}
Pauli Virtanen, Ralf Gommers, Travis~E. Oliphant, Matt Haberland, Tyler Reddy,
  David Cournapeau, Evgeni Burovski, Pearu Peterson, Warren Weckesser, Jonathan
  Bright, St{\'e}fan~J. {van der Walt}, Matthew Brett, Joshua Wilson, K.~Jarrod
  Millman, Nikolay Mayorov, Andrew R.~J. Nelson, Eric Jones, Robert Kern, Eric
  Larson, C~J Carey, {\.I}lhan Polat, Yu~Feng, Eric~W. Moore, Jake
  {VanderPlas}, Denis Laxalde, Josef Perktold, Robert Cimrman, Ian Henriksen,
  E.~A. Quintero, Charles~R. Harris, Anne~M. Archibald, Ant{\^o}nio~H. Ribeiro,
  Fabian Pedregosa, Paul {van Mulbregt}, and {SciPy 1.0 Contributors}. 2020.
\newblock \href {https://doi.org/10.1038/s41592-019-0686-2} {{{SciPy} 1.0:
  Fundamental Algorithms for Scientific Computing in Python}}.
\newblock \emph{Nature Methods}, 17:261--272.

\bibitem[{Vukovi{\'{c}} et~al.(2022)Vukovi{\'{c}}, Arora, Chang, Spitz, and
  West}]{Vukovi__2022}
Vuk Vukovi{\'{c}}, Akhil Arora, Huan-Cheng Chang, Andreas Spitz, and Robert
  West. 2022.
\newblock \href {https://doi.org/10.1145/3477495.3531696} {Quote erat
  demonstrandum: A web interface for exploring the quotebank corpus}.
\newblock In \emph{Proceedings of the 45th International {ACM} {SIGIR}
  Conference on Research and Development in Information Retrieval}. {ACM}.

\bibitem[{Wang and Komatsuzaki(2021)}]{gpt-j}
Ben Wang and Aran Komatsuzaki. 2021.
\newblock {GPT-J-6B}: A 6 billion parameter autoregressive language model.

\bibitem[{Wenzek et~al.(2019)Wenzek, Lachaux, Conneau, Chaudhary, Guzmán,
  Joulin, and Grave}]{https://doi.org/10.48550/arxiv.1911.00359}
Guillaume Wenzek, Marie-Anne Lachaux, Alexis Conneau, Vishrav Chaudhary,
  Francisco Guzmán, Armand Joulin, and Edouard Grave. 2019.
\newblock \href {https://doi.org/10.48550/ARXIV.1911.00359} {Ccnet: Extracting
  high quality monolingual datasets from web crawl data}.

\bibitem[{{W}es {M}c{K}inney(2010)}]{pandas-paper}
{W}es {M}c{K}inney. 2010.
\newblock \href {https://doi.org/10.25080/Majora-92bf1922-00a} {{D}ata
  {S}tructures for {S}tatistical {C}omputing in {P}ython}.
\newblock In \emph{{P}roceedings of the 9th {P}ython in {S}cience
  {C}onference}, pages 56 -- 61.

\bibitem[{Yang et~al.(2017)Yang, Fang, and Lin}]{10.1145/3077136.3080721}
Peilin Yang, Hui Fang, and Jimmy Lin. 2017.
\newblock \href {https://doi.org/10.1145/3077136.3080721} {Anserini: Enabling
  the use of lucene for information retrieval research}.
\newblock In \emph{Proceedings of the 40th International ACM SIGIR Conference
  on Research and Development in Information Retrieval}, SIGIR '17, page
  1253–1256, New York, NY, USA. Association for Computing Machinery.

\bibitem[{Yang et~al.(2022)Yang, Chen, PourNejatian, Shin, Smith, Parisien,
  Compas, Martin, Flores, Zhang, Magoc, Harle, Lipori, Mitchell, Hogan,
  Shenkman, Bian, and Wu}]{https://doi.org/10.48550/arxiv.2203.03540}
Xi~Yang, Aokun Chen, Nima PourNejatian, Hoo~Chang Shin, Kaleb~E Smith,
  Christopher Parisien, Colin Compas, Cheryl Martin, Mona~G Flores, Ying Zhang,
  Tanja Magoc, Christopher~A Harle, Gloria Lipori, Duane~A Mitchell, William~R
  Hogan, Elizabeth~A Shenkman, Jiang Bian, and Yonghui Wu. 2022.
\newblock \href {https://doi.org/10.48550/ARXIV.2203.03540} {Gatortron: A large
  clinical language model to unlock patient information from unstructured
  electronic health records}.

\bibitem[{Zhang et~al.(2020)Zhang, Gupta, Tang, Han, Pradeep, Lu, Zhang,
  Nogueira, Cho, Fang, and Lin}]{zhang2020covidex}
Edwin Zhang, Nikhil Gupta, Raphael Tang, Xiao Han, Ronak Pradeep, Kuang Lu, Yue
  Zhang, Rodrigo Nogueira, Kyunghyun Cho, Hui Fang, and Jimmy Lin. 2020.
\newblock \href {https://doi.org/10.18653/v1/2020.sdp-1.5} {Covidex: Neural
  ranking models and keyword search infrastructure for the {COVID}-19 open
  research dataset}.
\newblock In \emph{Proceedings of the First Workshop on Scholarly Document
  Processing}, pages 31--41, Online. Association for Computational Linguistics.

\bibitem[{Zhang et~al.(2022)Zhang, Roller, Goyal, Artetxe, Chen, Chen, Dewan,
  Diab, Li, Lin, Mihaylov, Ott, Shleifer, Shuster, Simig, Koura, Sridhar, Wang,
  and Zettlemoyer}]{https://doi.org/10.48550/arxiv.2205.01068}
Susan Zhang, Stephen Roller, Naman Goyal, Mikel Artetxe, Moya Chen, Shuohui
  Chen, Christopher Dewan, Mona Diab, Xian Li, Xi~Victoria Lin, Todor Mihaylov,
  Myle Ott, Sam Shleifer, Kurt Shuster, Daniel Simig, Punit~Singh Koura, Anjali
  Sridhar, Tianlu Wang, and Luke Zettlemoyer. 2022.
\newblock \href {https://doi.org/10.48550/ARXIV.2205.01068} {Opt: Open
  pre-trained transformer language models}.

\end{thebibliography}
\bibliographystyle{acl_natbib}




\end{document}